\documentclass[cameraready]{Interspeech}

\title{Your Voice Cloning System is Secretly a Voice Anonymizer}

\author[affiliation={1}, ]{Romolo}{Muletta}
\author[affiliation={1}, orcid=0009-0002-2224-2417]{Felix}{Saaro}
\author[affiliation={1}, orcid=0009-0007-3059-8516]{Mark}{Cieliebak}
\author[affiliation={1}, orcid=0000-0002-8405-1344, correspondingauthor]{Jan}{Deriu}

\address{
    $^1$ Centre for Artificial Intelligence ZHAW School of Engineering 
}

\email{muletrom@zhaw.ch, saaf@zhaw.ch, ciel@zhaw.ch, deri@zhaw.ch}

\keywords{voice anonymization}

\usepackage{comment}
\usepackage{multirow}
\usepackage{relsize}

\begin{document}

\maketitle

\begin{abstract}
Speaker anonymization suppresses speaker-identifying attributes from speech while preserving linguistic content and quality. We propose repurposing XTTSv2, a multilingual voice cloning model trained on 27k hours of speech, for speaker anonymization without retraining. Our key insight is that XTTSv2's voice cloning capabilities preserve prosodic structure independently of speaker identity, enabling voice conversion by conditioning on a pseudo-speaker. We introduce an iterative refinement strategy that balances privacy and utility by maximizing a harmonic mean of speaker dissimilarity and intelligibility. Evaluated on seven European languages across CommonVoice and Multilingual LibriSpeech, our system achieves near-optimal privacy (EER $\approx$ 0.49), competitive intelligibility, and substantially better speech quality than dedicated anonymization baselines, while requiring no language-specific training. We release the code here~\url{https://github.com/rm00cr/coqui-tts}.
\end{abstract}

\section{Introduction}
Speech signals contain biometric information that can be used to identify speakers, which can be misused for unauthorized profiling. The task of speaker anonymization is to automatically suppress speaker identifying attributes from the speech, making identification impossible~\cite{tomashenko2020voiceprivacy}. That is, to modify speech so that the speaker's identity remains hidden while preserving linguistic content, prosody, and speech quality. The original speaker's voice is converted to an artificial voice, so that the output cannot be linked to the original speaker by an automatic speaker verification (ASV) system. The VoicePrivacy Challenge~\cite{tomashenko2020voiceprivacy, panariello2024voiceprivacy, tomashenko2024voiceprivacy} has emerged as the primary benchmark for evaluating speaker anonymization systems. The challenge defines anonymization as a privacy-utility trade-off: systems must maximize the equal error rate (EER) of automatic speaker verification (ASV) attacks while minimizing degradation to word error rate (WER) and preserving emotional content. 

Speaker anonymization approaches can be broadly categorized into signal processing methods and neural network-based methods. Signal processing techniques, such as the McAdams coefficient~\cite{patino21mcadams} or phase vocoder-based time-scale modification~\cite{mawalim22_spsc}, offer parameter-free anonymization but provide limited privacy protection against sophisticated attackers. Recent work is based on neural network anonymization strategies. Meyer et al.~\cite{meyer2023gan} demonstrated that Generative Adversarial Networks (GANs) can generate diverse pseudo-speaker embeddings with controllable dissimilarity to source speakers. Meyer et al.~\cite{meyer2023prosody} further showed that prosody can be preserved independently of speaker identity by cloning pitch, energy, and duration patterns while anonymizing the speaker embedding. Lv et al.~\cite{lv2023salt} proposed SALT, which leverages self-supervised learning (WavLM) to extract latent features and creates pseudo-speakers using multiple speaker representations in latent space, achieving state-of-the-art speaker distinctiveness while maintaining privacy. Neural audio codecs (NAC) language models~\cite{panariello2024speaker} exploit the natural disentanglement between semantic tokens (encoding linguistic content) and acoustic tokens (encoding speaker characteristics) to achieve strong anonymization while maintaining speech quality.

Despite significant progress, speaker anonymization research has been predominantly limited to English. This linguistic bias severely constrains the applicability of anonymization technologies to the vast majority of the world's languages and speakers. Only recently have researchers begun addressing multilingual scenarios. Miao et al.~\cite{miao22_odyssey,miao22_interspeech} proposed a language-independent approach using self-supervised learning (SSL) models such as HuBERT to extract content representations without explicit language-specific components. Their system, evaluated on English and Mandarin, achieved promising results but still required careful domain adaptation for optimal performance. Meyer et al.~\cite{meyer24_interspeech} extended their GAN-based anonymization system to nine languages by replacing monolingual ASR and text-to-speech (TTS) components with multilingual counterparts. Their results demonstrated that speaker embeddings trained on English generalize well across languages, and that anonymization performance is primarily determined by the quality of the language-specific synthesis component. This finding suggests that high-quality multilingual TTS systems could enable effective anonymization across diverse languages.

State-of-the-art systems typically require training complex multi-component pipelines from scratch, including speaker encoders, linguistic feature extractors, and neural vocoders. This complexity creates barriers to entry for low-resource languages where training data for each component may be scarce.

In parallel, the text-to-speech (TTS) community has developed powerful multilingual voice cloning models. XTTSv2~\cite{casanova24_interspeech} achieves state-of-the-art zero-shot voice cloning across 16 languages with only a few seconds of reference audio. XTTS was trained on 27,000 hours of multilingual speech and enables cross-lingual voice cloning while preserving prosody and naturalness.

In this work, we observe that voice cloning and speaker anonymization share a common core operation: voice conversion conditioned on a target speaker identity. Rather than training a dedicated anonymization system from scratch, we propose repurposing XTTSv2 for speaker anonymization by leveraging its inference-time flexibility. Our key insight is that XTTSv2 naturally preserves prosodic structure independently of speaker identity by leveraging its voice cloning capabilities. By combining the codebook representation used in XTTS with a synthesized pseudo-speaker identity, we can transform the original speaker's voice while maintaining prosody, linguistic content, and speech quality. Our contributions are twofold:

\begin{itemize}
\item We introduce a simple method for repurposing XTTSv2, a multilingual voice-cloning model, for speaker anonymization without retraining.
\item We propose an iterative refinement strategy that balances privacy and utility by maximizing a harmonic mean of speaker dissimilarity and intelligibility preservation.
\end{itemize}

Unlike previous work that builds complex systems from scratch, our approach leverages the capabilites state-of-the-art TTS model. This makes speaker anonymization more accessible, particularly for languages already supported by XTTSv2, while achieving competitive privacy-utility trade-offs. We will release the code and the generated, anonymized audio files. For the version under review, we add the fully functional annotation tool and anonymized audio files as supplemental material. 

\section{Model}

\begin{figure}[!t]
    \centering
    \includegraphics[width=0.75\linewidth]{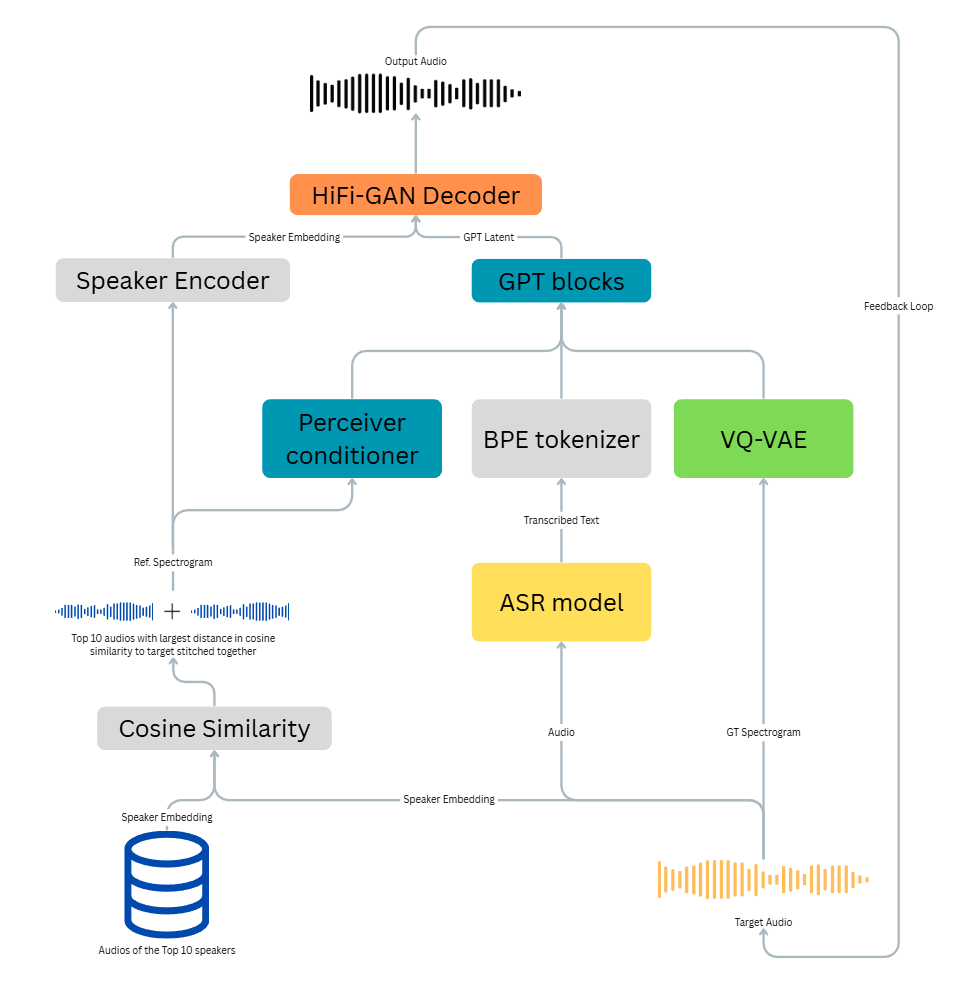}
    \caption{Speaker anonymization pipeline using XTTSv2. The original speech is encoded into a codebook representation via VQ-VAE, preserving prosody. The GPT-2 backbone transforms this representation to match the pseudo-speaker identity provided by the Perceiver conditioner.}
    \label{fig:architecture}
\end{figure}

We repurpose XTTSv2~\cite{casanova24_interspeech}, a multilingual voice cloning model trained on 27k hours of speech, for speaker anonymization. Rather than training a dedicated system, we exploit the model's voice conversion capabilities by conditioning generation on both a target speaker identity and the prosodic structure of the original speech.

\subsection{XTTSv2 Components}

We briefly describe the relevant components of XTTSv2; see Figure~\ref{fig:architecture}.

\noindent\textbf{VQ-VAE:} A Vector Quantized Variational Autoencoder~\cite{betker2023better} that encodes mel-spectrograms into discrete codebook indices. In standard voice cloning, this component is used only during training. For anonymization, we use it at inference time to extract a codebook representation of the original speech that preserves prosodic structure (pitch contour, rhythm, energy) independently of speaker identity.

\noindent\textbf{Perceiver Conditioner:} Six attention layers followed by a Perceiver Resampler~\cite{alayrac2022flamingo}, which compresses variable-length reference audio into a fixed set of 32 embeddings representing the target speaker's voice characteristics.

\noindent\textbf{GPT-2 Backbone:} A decoder-only transformer~\cite{radford2019gpt2} that autoregressively predicts codebook tokens. In standard cloning, it receives only the Perceiver output and text tokens. For anonymization, we additionally provide the original speech's codebook representation, enabling the model to preserve prosody while transforming speaker identity.

\noindent\textbf{Text Encoder:} A BPE tokenizer~\cite{gage1994compression} with 6681 tokens, providing the linguistic content.

\noindent\textbf{HiFi-GAN Vocoder:} Synthesizes the final waveform from the transformer's latent output, additionally conditioned on a speaker embedding from the H/ASP model~\cite{heo2020clova}.

\subsection{Anonymization Pipeline}
The full pipeline is depicted in Figure~\ref{fig:architecture}. It comprises an ASR model (in our case, Whisper-Large-V3~\cite{radford2023whisper}), a gender classifier, a pool of 10 reference speakers per language and gender, each with a set of utterances used to create a novel, anonymized voice, and the XTTSv2 model. Given speech to be anonymized:

\begin{enumerate}
    \item \textbf{Feature extraction:} Transcribe the speech using Whisper-Large-V3~\cite{radford2023whisper}, extract the codebook representation via VQ-VAE, and compute an ECAPA2~~\cite{thienpondt2023ecapa2} speaker embedding.
    
    \item \textbf{Pseudo-speaker construction:} Classify the speaker's gender and select the corresponding reference pool. For each of the 10 pool speakers, select the utterance whose ECAPA2 embedding has the lowest cosine similarity to the original speaker's embedding. Concatenate these 10 utterances and pass them through the Perceiver conditioner to obtain a composite pseudo-speaker representation. At the same time, they are passed through the speaker encoder to create a speaker embedding. 
    
    \item \textbf{Voice conversion:} Concatenate the embeddings from the perceiver conditioner, text tokens, and original codebook representation. The GPT-2 backbone transforms this into a new codebook sequence aligned with the pseudo-speaker identity.
    
    \item \textbf{Synthesis:} The HiFi-GAN vocoder generates the anonymized waveform from the transformed codebook and the pseudo-speaker embedding.
\end{enumerate}

\subsection{Iterative Refinement}

We observed that the transformer occasionally reverts to characteristics of the original speaker mid-utterance, likely due to attention over the original codebook tokens. To address this, we iteratively reapply the pipeline to its own output until convergence. At each iteration, we compute a selection criterion balancing privacy and utility:
\begin{equation}
    H = \frac{2 \cdot (1 - \text{WER}) \cdot (1 - \text{Sim}_{\text{orig}})}{(1 - \text{WER}) + (1 - \text{Sim}_{\text{orig}})}
    \label{eq:harmonic_mean}
\end{equation}
where $\text{Sim}_{\text{orig}}$ is the cosine similarity between the anonymized and original speaker embeddings. We select the iteration that maximizes $H$.

\subsection{Pseudo-Speaker Pool Construction}
\label{sec:pseudo_speakers}
The quality of anonymized speech depends heavily on the reference audio used to define the pseudo-speaker. We construct language- and gender-specific pools of 10 high-quality reference speakers from Multilingual LibriSpeech (MLS)~\cite{pratap2020MLSAL}, selected as follows:

For each candidate speaker, we anonymize a held-out set of utterances using that speaker as the target voice. We then evaluate each candidate on two criteria: (1) speaker dissimilarity, measured as cosine distance between ECAPA2~\cite{thienpondt2023ecapa2} embeddings of the original and anonymized speech, and (2) intelligibility preservation, measured as WER of the anonymized speech. We retain the 10 speakers per language and gender who achieve the best trade-off between these criteria.

\begin{table}[t!]
\centering
\small
\resizebox{0.47\textwidth}{!}{%
\begin{tabular}{lrrrr}
\toprule
& \multicolumn{2}{c}{\textbf{Common Voice}} & \multicolumn{2}{c}{\textbf{MLS}} \\
\cmidrule(lr){2-3} \cmidrule(lr){4-5}
\textbf{Language} & Spkrs & Samples & Spkrs & Samples \\
\midrule
German (de)     & 819   & 3,003  & 60  & 6,863 \\
English (en)    & 2,538 & 4,481  & 84  & 7,576 \\
Spanish (es)    & 2,068 & 6,429  & 40  & 4,792 \\
French (fr)     & 934   & 3,506  & 36  & 4,838 \\
Italian (it)    & 1,083 & 6,836  & 20  & 2,510 \\
Dutch (nl)      & 631   & 12,400 & 12  & 6,166 \\
Portuguese (pt) & 711   & 7,224  & 20  & 1,697 \\
\midrule
\textbf{Total}  & \textbf{8,784} & \textbf{43,879} & \textbf{272} & \textbf{34,442} \\
\bottomrule
\end{tabular}

}
\caption{Evaluation data: speaker and sample counts across seven languages (dev + test splits combined).}
\label{tab:data_overview}
\end{table}

\begin{table*}[t!]
\centering
\vspace{0.5em}
\setlength{\tabcolsep}{5pt}
\small
\begin{tabular}{ll cccccccc cccccccc c}
\toprule
& & \multicolumn{8}{c}{\textbf{EER} $\uparrow$} & \multicolumn{8}{c}{\textbf{WER} $\downarrow$} & \textbf{$\Delta$UTM. $\uparrow$} \\
\cmidrule(lr){3-10} \cmidrule(lr){11-18} \cmidrule(lr){19-19}
& \textbf{Model} & \textbf{de} & \textbf{en} & \textbf{es} & \textbf{fr} & \textbf{it} & \textbf{nl} & \textbf{pt} & \textbf{Avg} & \textbf{de} & \textbf{en} & \textbf{es} & \textbf{fr} & \textbf{it} & \textbf{nl} & \textbf{pt} & \textbf{Avg} & \textbf{Avg} \\
\midrule
\multirow{3}{*}{\rotatebox[origin=c]{90}{\smaller\textbf{CV}}}
& SALT         & .37 & .35 & .38 & .38 & .37 & .34 & .38 & .37 & .21 & .37 & .18 & .32 & .23 & .17 & .35 & .26 & -0.74 \\
& MultiLingual & .46 & .46 & .46 & .46 & .46 & .46 & .47 & .46 & .20 & .31 & .19 & .28 & .25 & .23 & .39 & .27 & -0.62 \\
& XTTSv2 (Ours) & \textbf{.49} & \textbf{.49} & \textbf{.49} & \textbf{.50} & \textbf{.49} & \textbf{.50} & \textbf{.48} & \textbf{.49} & \textbf{.16} & \textbf{.28} & \textbf{.10} & \textbf{.20} & \textbf{.12} & \textbf{.07} & \textbf{.22} & \textbf{.16} & \textbf{+0.17} \\
\midrule
\multirow{3}{*}{\rotatebox[origin=c]{90}{\smaller\textbf{MLS}}}
& SALT         & .35 & .29 & .32 & .35 & .27 & .16 & .36 & .30 & \textbf{.09} & \textbf{.12} & \textbf{.06} & \textbf{.12} & \textbf{.17} & \textbf{.23} & \textbf{.13} & \textbf{.13} & -1.29 \\
& MultiLingual & .46 & .45 & .45 & .45 & \textbf{.44} & .36 & \textbf{.45} & .44 & .14 & .17 & .08 & .17 & .20 & .24 & .21 & .17 & -1.23 \\
& XTTSv2 (Ours) & \textbf{.49} & \textbf{.48} & \textbf{.48} & \textbf{.48} & .42 & \textbf{.44} & .44 & \textbf{.46} & .15 & .13 & .08 & \textbf{.12} & \textbf{.17} & .24 & .20 & .16 & \textbf{-0.35} \\
\bottomrule
\end{tabular}
\caption{Speaker anonymization results across seven languages (dev+test combined). EER: higher is better (0.5 = optimal privacy). WER: lower is better. $\Delta$UTMOS: naturalness change from original. Best results in \textbf{bold}.}
\label{tab:full_results}
\end{table*}

\section{Experimental Setup}

\subsection{Datasets and Languages}
We evaluate speaker anonymization on two multilingual datasets: Multilingual LibriSpeech (MLS)~\cite{pratap2020MLSAL} and CommonVoice 23.0 (CV)~\cite{ardila-etal-2020-common}. We focus on seven languages: English (en), German (de), French (fr), Spanish (es), Italian (it), Portuguese (pt), and Dutch (nl). Table~\ref{tab:data_overview} shows the test-set overview.

For MLS, we use the full dev and test splits provided by the dataset. The MLS dev and test splits contain 272 speakers with high-quality audiobook recordings. For CV, we use the dev and test splits containing 8,784 speakers. Generally, MLS samples exhibit higher audio quality than CV samples, which often contain background noise and are recorded using consumer-grade laptop or smartphone microphones. This quality difference allows us to evaluate robustness across varying acoustic conditions.

The pseudo-speaker pools are constructed from MLS training splits, which provide high-quality, clean speech necessary for effective voice synthesis. We use 10 speakers per language-gender combination, selected based on their anonymization effectiveness, as described in Section~\ref{sec:pseudo_speakers}.

\subsection{Baseline Systems}

We compare our approach against two competitive anonymization systems:

\noindent\textbf{SALT~\cite{lv2023salt}} uses self-supervised learning features from WavLM to extract latent representations. Pseudo-speakers are created through interpolation and extrapolation of multiple speaker representations in latent space. SALT achieved first place in the VoicePrivacy Challenge 2022, demonstrating state-of-the-art speaker distinctiveness while maintaining privacy. We use the official implementation\footnote{\url{https://github.com/BakerBunker/SALT}} with default parameters.

\noindent\textbf{MultiLingual~\cite{meyer24_interspeech}} extends GAN-based speaker anonymization to nine languages using multilingual ASR (Whisper-large-v3) and TTS (IMS Toucan) components. The system extracts linguistic content, prosody, and speaker embeddings, then replaces the speaker embedding with a GAN-generated pseudo-speaker while preserving prosodic features. We use the official implementation\footnote{\url{https://github.com/DigitalPhonetics/speaker-anonymization}} configured for the seven evaluation languages.

\subsection{Our Anonymization System}

Our system repurposes XTTSv2~\cite{casanova24_interspeech} for speaker anonymization without retraining. The pipeline consists of the following components:

\noindent\textbf{Feature Extraction:}
\begin{itemize}
\item \textbf{Transcription:} Whisper-Large-V3~\cite{radford2023whisper} for multilingual automatic speech recognition
\item \textbf{Prosody:} VQ-VAE codebook representation extracted from XTTSv2's encoder
\item \textbf{Speaker Identity:} ECAPA2~\cite{thienpondt2023ecapa2} embeddings for both original and pseudo-speaker characterization
\item \textbf{Gender:} Binary gender classifier for pool selection
\end{itemize}

\noindent\textbf{Pseudo-Speaker Construction:}
For each utterance, we:
\begin{enumerate}
\item Classify the speaker's gender
\item Select the corresponding reference pool (10 speakers per language/gender)
\item For each pool speaker, choose the utterance with the lowest ECAPA2 cosine similarity to the original speaker
\item Concatenate these 10 utterances and process through XTTSv2's Perceiver conditioner to create a composite pseudo-speaker representation
\end{enumerate}

\noindent\textbf{Voice Conversion:}
The GPT-2 backbone in XTTSv2 receives:
\begin{itemize}
\item Pseudo-speaker embeddings from the Perceiver conditioner
\item Text tokens from the Whisper transcription
\item Original codebook representation preserving prosodic structure
\end{itemize}
The HiFi-GAN vocoder~\cite{kong2020hifigan} synthesizes the final anonymized waveform conditioned on the transformed codebook and pseudo-speaker embedding.

\noindent\textbf{Iterative Refinement:}
We apply the anonymization pipeline iteratively to its own output, computing, at each iteration, a harmonic mean that balances privacy and utility, as described in Equation~\ref{eq:harmonic_mean}.

\subsection{Evaluation Protocol}

\noindent\textbf{Privacy Evaluation:}
Privacy is measured using Equal Error Rate (EER) from an automatic speaker verification (ASV) system. We use an ECAPA2 as our ASV system. Higher EER indicates better anonymization; 50\% corresponds to random-guess performance (perfect anonymization).

\noindent\textbf{Intelligibility:} Computed using Whisper-Large-V3 on the anonymized speech, and then computed the WER between the original transcript and the generated transcript of the anonymized speech. Lower WER indicates better intelligibility preservation.

\noindent\textbf{Speech Quality:}  We measure speech quality using the UTokyo-SaruLab MOS Prediction System (UTMOS)~\cite{saeki2022utmos}. We report $\Delta$UTMOS = UTMOS$_{\text{anonymized}}$ - UTMOS$_{\text{original}}$, where values near 0 indicate preserved quality, and higher values indicate better relative quality (range: -4 to +4).

\noindent\textbf{Comparative Mean Opinion Score (CMOS):}
We conduct human listening tests on English to evaluate perceived speech quality. We randomly sample 30 utterances from the MLS English test set and anonymize them using all three systems. Nine fluent English speakers rate each original-anonymized pair on a 7-point scale from -3 (much worse) to +3 (much better), with 0 indicating equal quality. Each system receives 270 ratings (9 raters × 30 samples).

\subsection{Implementation Details}
All experiments use the XTTSv2 version from the Coqui TTS library~\footnote{https://github.com/coqui-ai/TTS}. For fair comparison, all systems use the same evaluation models (Whisper-Large-V3 for WER, ECAPA2 for EER).

\section{Results}

Table~\ref{tab:full_results} presents the main results comparing our XTTSv2-based system against SALT and MultiLingual baselines across seven languages on both CommonVoice (CV) and Multilingual LibriSpeech (MLS) datasets.

\noindent\textbf{Privacy Protection (EER): } Our system achieves an average EER of 0.49 on CV and 0.46 on MLS, approaching the theoretical maximum of 0.50 (indicating random guessing by the ASV system). This represents improved anonymization compared to SALT (0.37 CV, 0.30 MLS) and performs comparably or better than MultiLingual (0.46 CV, 0.44 MLS). On CV, EER values are consistent across languages (0.48--0.50), while on MLS, they range from 0.42 to 0.49. The lower performance on some MLS languages may reflect differences in recording quality or speaker diversity within the dataset.

\noindent\textbf{Intelligibility (WER):} On CV, our approach achieves an average WER of 0.16, compared to 0.26 for SALT and 0.27 for MultiLingual. This corresponds to a 38\% relative reduction compared to the baselines. Individual language WER ranges from 0.07 (Dutch) to 0.28 (English). On MLS, SALT achieves the lowest average WER at 0.13, while our system obtains 0.16, slightly outperforming MultiLingual (0.17). 

\noindent\textbf{Speech Quality ($\Delta$UTMOS):} On CV, our system achieves $\Delta$UTMOS of +0.17, indicating improved quality relative to the original recordings. SALT and MultiLingual show degradation of -0.74 and -0.62, respectively. The positive value for our system likely reflects the cleaner synthesis, which compensates for noise in the original CV recordings. On MLS, all systems show quality degradation because the audiobook recordings have a high baseline quality. Our system achieves $\Delta$UTMOS of -0.35, compared to -1.29 for SALT and -1.23 for MultiLingual.

\begin{table}[t!]
\centering
\small
\begin{tabular}{lcc}
\toprule
\textbf{System} & \textbf{CMOS} & \textbf{Std Dev} \\
\midrule
SALT & $-1.00$ & $\pm1.12$ \\
MultiLingual & $-1.85$ & $\pm0.88$ \\
XTTSv2 (Ours) & $\mathbf{-0.90}$ & $\pm1.39$ \\
\bottomrule
\end{tabular}
\caption{Comparative Mean Opinion Score (CMOS) on English MLS test samples (30 samples, 9 raters, 270 total comparisons per system). Anonymized speech compared against original recordings. Scale: -3 (much worse) to +3 (much better), 0 (equal quality). Higher values indicate better quality preservation.}
\label{tab:cmos}
\end{table}
\noindent\textbf{Naturalness (CMOS):} Table~\ref{tab:cmos} shows human evaluation results for English on MLS. All systems exhibit quality degradation relative to the high-quality audiobook originals, consistent with the negative $\Delta$UTMOS values observed on MLS. Our system achieves the least degradation (CMOS = -0.90), outperforming SALT (-1.00) and MultiLingual (-1.85). The results confirm that XTTSv2's vocoder better preserves naturalness despite anonymization.

\noindent\textbf{Iterative Refinement.} 
The iterative refinement strategy selects different iterations across utterances: iteration 0 is chosen in 19.6\% of cases, iterations 1-3 are each selected approximately 18\% of the time, and iteration 4 is chosen in 25.0\% of cases. This distribution indicates that different utterances require different levels of refinement. None of the manually tested audios exhibit the mid-utterance degradation that motivated the iterative refinement. 

\section{Discussion \& Conclusion}
We presented a speaker anonymization approach that leverages XTTSv2 for privacy protection without retraining, achieving near-optimal EER across seven languages while substantially improving speech quality relative to dedicated baselines. Rather than building anonymization systems from scratch, our work demonstrates that the community can leverage existing multilingual TTS models. Future work should address adversarial robustness against informed attackers, extend to non-European languages, and explore generalization to other voice cloning architectures.

\section{AI Declaration}
We used Claude Opus 4.5 to edit and rewrite passages of text for better readability. 

\bibliographystyle{IEEEtran}
\bibliography{mybib}

\end{document}